\documentclass[a4paper,oneside,onecolumn]{IEEEtran}

\usepackage{epsfig,latexsym,cite,graphicx,amsmath,amssymb,amsfonts,multirow,booktabs,color,soul,overpic,algorithmic,float,threeparttable,makecell,graphicx,hhline}
\usepackage[linesnumbered,ruled,vlined]{algorithm2e}
\usepackage[caption=false,font=footnotesize]{subfig}
\usepackage[top=1in,bottom=1in,left=1in,right=1in]{geometry}

\definecolor{hl}{rgb}{0.75,0.75,0.75} % Highlight for table
\sethlcolor{hl}
\newcommand{\NAME}[1]{PlatEMO#1}
\newcommand{\MR}[2]{\multirow{#1}{*}{#2}}

\begin{document}

\title{\NAME{}: A MATLAB Platform for Evolutionary Multi-Objective Optimization}

\author{
Ye Tian$^1$, Ran Cheng$^2$, Xingyi Zhang$^1$, and Yaochu Jin$^3$\\
$^1$School of Computer Science and Technology, Anhui University, Hefei, 230601, China\\
$^2$School of Computer Science, University of Birmingham, Birmingham, B15 2TT, U.K.\\
$^3$Department of Computer Science, University of Surrey, Guildford, GU2 7XH, U.K.\\
\thanks{Corresponding author: Xingyi Zhang (E-mail: xyzhanghust@gmail.com)}
}

\date{}
\maketitle

\begin{abstract}
Over the last three decades, a large number of evolutionary algorithms have been developed for solving multi-objective optimization problems. However, there lacks an up-to-date and comprehensive software platform for researchers to properly benchmark existing algorithms and for practitioners to apply selected algorithms to solve their real-world problems. The demand of such a common tool becomes even more urgent, when the source code of many proposed algorithms has not been made publicly available. To address these issues, we have developed a MATLAB platform for evolutionary multi-objective optimization in this paper, called PlatEMO, which includes more than 50 multi-objective evolutionary algorithms and more than 100 multi-objective test problems, along with several widely used performance indicators. With a user-friendly graphical user interface, PlatEMO enables users to easily compare several evolutionary algorithms at one time and collect statistical results in Excel or LaTeX files.
More importantly, PlatEMO is completely open source, such that users are able to develop new algorithms on the basis of it.
This paper introduces the main features of PlatEMO and illustrates how to use it for performing comparative experiments, embedding new algorithms, creating new test problems, and developing performance indicators.
Source code of PlatEMO is now available at: http://bimk.ahu.edu.cn/index.php?s=/Index/Software/index.html.
\end{abstract}

\def\IEEEkeywordsname{Keywords}

\section{Introduction}

Multi-objective optimization problems (MOPs) widely exist in computer science such as data mining \cite{mukhopadhyay2014survey}, pattern recognition \cite{handl2007evolutionary}, image processing \cite{lazzerini2010multi} and neural network \cite{pettersson2007genetic}, as well as many other application fields~\cite{app-bio1,app-eco,app-eng1,app-eng2}.
An MOP consists of two or more conflicting objectives to be optimized, and there often exist a set of optimal solutions trading off between different objectives. Since the vector evaluated genetic algorithm (VEGA) was proposed by Schaffer in 1985 \cite{VEGA}, a number of multi-objective evolutionary algorithms (MOEAs) have been proposed and shown their superiority in tackling MOPs during the last three decades.
For example, several MOEAs based on Pareto ranking selection and fitness sharing mechanism including multi-objective genetic algorithm (MOGA) \cite{MOGA}, non-dominated sorting genetic algorithm (NSGA) \cite{NSGA}, and niched Pareto genetic algorithm (NPGA) \cite{NPGA} were proposed in the 1990s.
From 1999 to 2002, some MOEAs characterized by the elitism strategy were developed, such as non-dominated sorting genetic algorithm II (NSGA-II) \cite{NSGA-II}, strength Pareto evolutionary algorithm 2 (SPEA2) \cite{SPEA2}, Pareto envelope-based selection algorithm II (PESA-II) \cite{PESA-II} and cellular multiobjective genetic algorithm (cellular MOGA) \cite{cell-ea}.
Afterwards, the evolutionary algorithm based on decomposition (MOEA/D) was proposed in 2007 \cite{MOEAd}, and some other MOEAs following the basic idea of MOEA/D had also been developed since then \cite{moead-de,MOEAD-M2M,MOEAD-DU,Ishibuchi2016performance}.

In spite of the large number of MOEAs in the literature \cite{zhou2011multiobjective}, there often exist some difficulties in applying and using these algorithms since the source code of most algorithms had not been provided by the authors.
Besides, it is also difficult to make benchmark comparisons between MOEAs due to the lack of a general experimental environment.
To address such issues, several MOEA libraries have been proposed \cite{framwork-jMetal,framework-Opt4J,framework-OTL,framework-ParadisEO,framework-PISA} to provide uniform experimental environments for users, which have greatly advanced the multi-objective optimization research and the implementation of new ideas.
For example, the C-based multi-objective optimization library PISA~\cite{framework-PISA}\footnote{PISA: http://www.tik.ee.ethz.ch/pisa} separates the implementation into two components, i.e., the problem-specific part containing MOPs and operators, and the problem-independent part containing MOEAs.
These two components are connected by a text file-based interface in PISA.
jMetal \cite{framwork-jMetal}\footnote{jMetal: http://jmetal.sourceforge.net/index.html} is an object-oriented Java-based multi-objective optimization library consisting of various MOEAs and MOPs.
MOEA Framework\footnote{MOEA Framework: http://moeaframework.org/index.html} is another free and open source Java framework for multi-objective optimization, which provides a comprehensive collection of MOEAs and tools necessary to rapidly design, develop, execute and test MOEAs.
OTL \cite{framework-OTL}\footnote{OTL: http://github.com/O-T-L/OTL}, a C++ template library for multi-objective optimization, is characterized by object-oriented architecture, template technique, ready-to-use modules, automatically performed batch experiments and parallel computing.
Besides, a Python-based experimental platform has also been proposed as the supplement of OTL, for improving the development efficiency and performing batch experiments more conveniently.

\begin{table*}[htbp]
\renewcommand{\arraystretch}{0.7}
\caption{The 50 Multi-Objective Optimization Algorithms Included in the Current Version of \NAME{}.}
\begin{center}
\begin{tabular}{|c|c|c|}
\hline
\MR{2}{Algorithm}&Year of &\MR{2}{Description}\\
&Publication&\\
\hhline{|===|}
\multicolumn{3}{|c|}{Multi-Objective Genetic Algorithms}\\
\hline
SPEA2 \cite{SPEA2}&2001&Strength Pareto evolutionary algorithm 2\\
\hline
PSEA-II \cite{PESA-II}&2001&Pareto envelope-based selection algorithm II\\
\hline
NSGA-II \cite{NSGA-II}&2002&Non-dominated sorting genetic algorithm II\\
\hline
$\epsilon$-MOEA \cite{eMOEA}&2003&Multi-objective evolutionary algorithm based on $\epsilon$-dominance\\
\hline
IBEA \cite{IBEA}&2004&Indicator-based evolutionary algorithm\\
\hline
MOEA/D \cite{MOEAd}&2007&Multi-objective evolutionary algorithm based on decomposition\\
\hline
SMS-EMOA \cite{SMS}&2007&S metric selection evolutionary multi-objective optimization algorithm\\
\hline
MSOPS-II \cite{MSOPS-II}&2007&Multiple single objective Pareto sampling algorithm II\\
\hline
MTS \cite{tseng2009multiple}&2009&Multiple trajectory search\\
\hline
AGE-II \cite{AGE-II}&2013&Approximation-guided evolutionary algorithm II\\
\hline
NSLS \cite{NSLS}&2015&Non-dominated sorting and local search\\
\hline
BCE-IBEA \cite{BCE}&2015&Bi-criterion evolution for IBEA\\
\hline
\MR{2}{MOEA/IGD-NS \cite{IGD-NS}}&\MR{2}{2016}&Multi-objective evolutionary algorithm based on an\\
                                             &&enhanced inverted generational distance metric\\
\hhline{|===|}
\multicolumn{3}{|c|}{Many-Objective Genetic Algorithms}\\
\hline
HypE \cite{HypE}&2011&Hypervolume-based estimation algorithm\\
\hline
PICEA-g \cite{PICEA}&2013&Preference-inspired coevolutionary algorithm with goals\\
\hline
GrEA \cite{GrEA}&2013&Grid-based evolutionary algorithm\\
\hline
NSGA-III \cite{NSGA-III}&2014&Non-dominated sorting genetic algorithm III\\
\hline
A-NSGA-III \cite{NSGA-III2}&2014&Adaptive NSGA-III\\
\hline
SPEA2+SDE \cite{SDE}&2014&SPEA2 with shift-based density estimation\\
\hline
BiGE \cite{BiGE}&2015&Bi-goal evolution\\
\hline
EFR-RR \cite{MOEAD-DU}&2015&Ensemble fitness ranking with ranking restriction\\
\hline
I-DBEA \cite{I-DBEA}&2015&Improved decomposition based evolutionary algorithm\\
\hline
KnEA \cite{KnEA}&2015&Knee point driven evolutionary algorithm\\
\hline
\MR{2}{MaOEA-DDFC \cite{cheng2015many}}&\MR{2}{2015}&Many-objective evolutionary algorithm based on directional\\
                                                   &&diversity and favorable convergence\\
\hline
MOEA/DD \cite{MOEA-DD}&2015&Multi-objective evolutionary algorithm based on dominance and decomposition\\
\hline
MOMBI-II \cite{MOMBI-II}&2015&Many-objective metaheuristic based on the R2 indicator II\\
\hline
Two\_Arch2 \cite{Two_Arch2}&2015&Two-archive algorithm 2\\
\hline
\MR{2}{MaOEA-R\&D \cite{he2016many}}&\MR{2}{2016}&Many-objective evolutionary algorithm based on objective\\
                                                &&space reduction and diversity improvement\\
\hline
RPEA \cite{RPEA}&2016&Reference points-based evolutionary algorithm\\
\hline
RVEA \cite{RVEA}&2016&Reference vector guided evolutionary algorithm\\
\hline
RVEA* \cite{RVEA}&2016&RVEA embedded with the reference vector regeneration strategy\\
\hline
SPEA/R \cite{SPEA/R}&2016&Strength Pareto evolutionary algorithm based on reference direction\\
\hline
$\theta$-DEA \cite{tDEA}&2016&$\theta$-dominance based evolutionary algorithm\\

\hhline{|===|}
\multicolumn{3}{|c|}{Multi-Objective Genetic Algorithms for Large-Scale Optimization}\\
\hline
MOEA/DVA \cite{MOEA-DVA}&2016&Multi-objective evolutionary algorithm based on decision variable analyses\\
\hline
LMEA \cite{LMEA}&2016&Large-scale many-objective evolutionary algorithm\\

\hhline{|===|}
\multicolumn{3}{|c|}{Multi-Objective Genetic Algorithms with Preference}\\
\hline
g-NSGA-II \cite{reference2}&2009&g-dominance based NSGA-II\\
\hline
r-NSGA-II \cite{r-dominance}&2010&r-dominance based NSGA-II\\
\hline
WV-MOEA-P \cite{WV-MOEA-P}&2016&Weight vector based multi-objective optimization algorithm with preference\\

\hhline{|===|}
\multicolumn{3}{|c|}{Multi-objective Differential Algorithms}\\
\hline
GDE3 \cite{GDE3}&2005&Generalized differential evolution 3\\
\hline
MOEA/D-DE \cite{moead-de}&2009&MOEA/D based on differential evolution\\

\hhline{|===|}
\multicolumn{3}{|c|}{Multi-objective Particle Swarm Optimization Algorithms}\\
\hline
MOPSO \cite{MOPSO}&2002&Multi-objective particle swarm optimization\\
\hline
SMPSO \cite{SMPSO}&2009&Speed-constrained multi-objective particle swarm optimization\\
\hline
dMOPSO \cite{dMOPSO}&2011&Decomposition-based particle swarm optimization\\
\hline

\hhline{|===|}
\multicolumn{3}{|c|}{Multi-objective Memetic Algorithms}\\
\hline
M-PAES \cite{M-PAES}&2000&Memetic algorithm based on Pareto archived evolution strategy\\

\hhline{|===|}
\multicolumn{3}{|c|}{Multi-objective Estimation of Distribution Algorithms}\\
\hline
MO-CMA \cite{MO-CMA}&2007&Multi-objective covariance matrix adaptation\\
\hline
RM-MEDA \cite{IGD1}&2008&Regularity model-based multi-objective estimation of distribution algorithm\\
\hline
IM-MOEA \cite{moea-im}&2015&Inverse modeling multi-objective evolutionary algorithm\\

\hhline{|===|}
\multicolumn{3}{|c|}{Surrogate Model Based Multi-objective Algorithms}\\
\hline
ParEGO \cite{ParEGO}&2005&Efficient global optimization for Pareto optimization\\
\hline
SMS-EGO \cite{SMS-EGO}&2008&S-metric-selection-based efficient global optimization\\
\hline
K-RVEA \cite{K-RVEA}&2016&Kriging assisted RVEA\\
\hline
\end{tabular}
\end{center}
\label{tab:MOEA}
\end{table*}

\begin{table*}[htbp]
\renewcommand{\arraystretch}{0.85}
\caption{The 110 Multi-Objective Optimization Problems Included in the Current Version of \NAME{}.}
\begin{center}
\begin{tabular}{|c|c|c|}
\hline
\MR{2}{Problem}&Year of &\MR{2}{Description}\\
&Publication&\\
\hline
\MR{2}{MOKP \cite{SPEA}}&\MR{2}{1999}&Multi-objective 0/1 knapsack problem and \\
&&behavior of MOEAs on this problem analyzed in \cite{Ishibuchi2015mokb}\\
\hline
ZDT1--ZDT6 \cite{ZDT}&2000&Multi-objective test problems\\
\hline
mQAP \cite{mQAP}&2003&Multi-objective quadratic assignment problem\\
\hline
DTLZ1--DTLZ9 \cite{DTLZ}&2005&Scalable multi-objective test problems\\
\hline
\MR{2}{WFG1--WFG9 \cite{WFG}}&\MR{2}{2006}&Scalable multi-objective test problems and \\
&&degenerate problem WFG3 analyzed in~\cite{Ishibuchi2015Pareto}\\
\hline
MONRP \cite{MONRP}&2007&Multi-objective next release problem\\
\hline
MOTSP \cite{non-dominated-better}&2007&Multi-objective traveling salesperson problem\\
\hline
Pareto-Box \cite{ParetoBox}&2007&Pareto-Box problem\\
\hline
\MR{2}{CF1--CF10 \cite{UF-test}}&\MR{2}{2008}&Constrained multi-objective test problems for the\\
                                            &&CEC 2009 special session and competition\\
\hline
F1--F10 for RM-MEDA \cite{IGD1}&2008&The test problems designed for RM-MEDA\\
\hline
\MR{2}{UF1--UF12 \cite{UF-test}}&\MR{2}{2008}&Unconstrained multi-objective test problems for the\\
                                            &&CEC 2009 special session and competition\\
\hline
F1--F9 for MOEA/D-DE \cite{moead-de}&2009&The test problems extended from \cite{okabe2004test} designed for MOEA/D-DE\\                                            \hline
C1\_DTLZ1, C2\_DTLL2, C3\_DTLZ4&\MR{2}{2014}&Constrained DTLZ and\\
IDTLZ1, IDTLZ2\cite{NSGA-III2}&&inverted DTLZ\\
\hline
F1--F7 for MOEA/D-M2M \cite{MOEAD-M2M}&2014&The test problems designed for MOEA/D-M2M\\
\hline
F1--F10 for IM-MOEA \cite{moea-im}&2015&The test problems designed for IM-MOEA\\
\hline
BT1--BT9 \cite{MOEAD-CMA}&2016&Multi-objective test problems with bias\\
\hline
LSMOP1--LSMOP9 \cite{LSMOP}&2016&Large-scale multi-objective test problems\\
\hline
\end{tabular}
\end{center}
\label{tab:MOP}
\end{table*}

\begin{figure}
  \centering
  \includegraphics[width=0.9\linewidth]{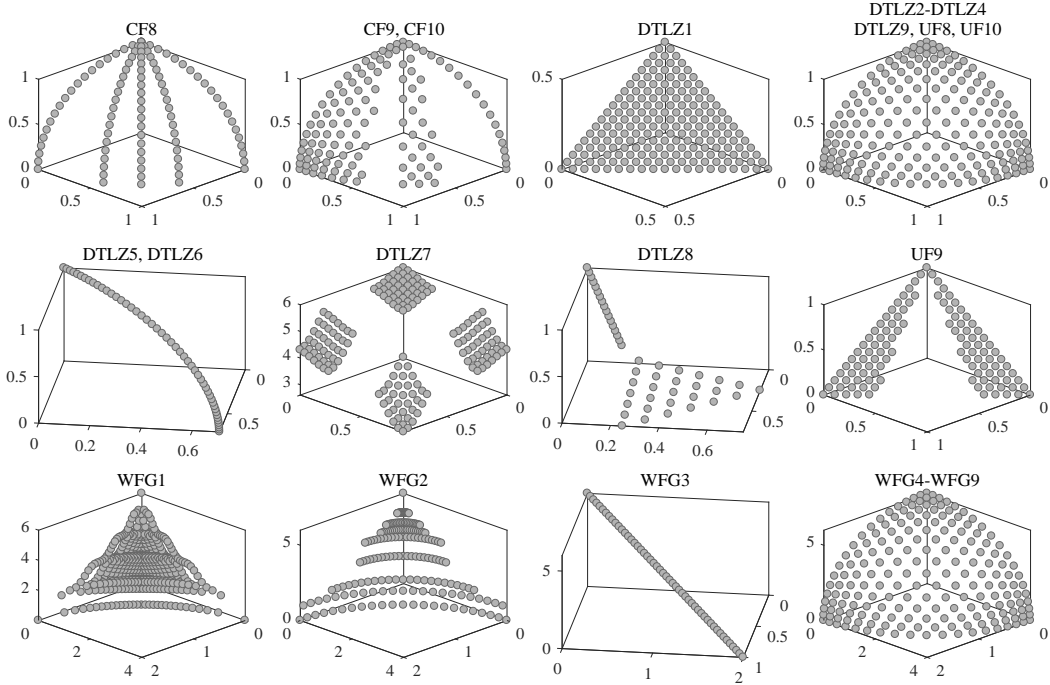}\\
  \caption{The reference points generated by \NAME{} on the Pareto fronts of CF, DTLZ, UF and WFG test suites with 3 objectives.}
  \label{fig:PF}
\end{figure}

It is encouraging that there are several MOEA libraries dedicated to the development of evolutionary multi-objective optimization (EMO), but unfortunately, most of them are still far from useful and practical to most researchers.
On one hand, the existing MOEA libraries cannot catch up with the development of MOEAs, where most of the MOEAs included in them are outdated and not able to cover the state-of-the-arts.
On the other hand, due to the lack of professional GUI for experimental settings and algorithmic configurations, these libraries are diffuclt to be used or extended, especially for beginners who are not familiar with EMO.
In order to collect more modern MOEAs and make the implementation of experiments on MOEAs easier, in this paper, we introduce a MATLAB-based EMO platform called \NAME{}\footnote{\NAME{}: http://bimk.ahu.edu.cn/index.php?s=/Index/Software/index.html}.
Compared to existing EMO platforms, \NAME{} has the following main advantages:
\begin{itemize}

\item{\emph{Rich Library.}} \NAME{} now includes 50 existing popular MOEAs as shown in Table \ref{tab:MOEA}, where most of them are representative algorithms published in top journals, including multi-objective genetic algorithms, multi-objective differential evolution algorithms, multi-objective particle swarm optimization algorithms, multi-objective memetic algorithms, multi-objective estimation of distribution algorithms, and so on.
\NAME{} also contains 110 MOPs from 16 popular test suites covering various difficulties, which are listed in Table \ref{tab:MOP}.
In addition, there are a lot of performance indicators provided by \NAME{} for experimental studies, including Coverage \cite{SPEA}, generational distance (GD) \cite{GD}, hypervolume (HV) \cite{HV-indicator}, inverted generational distance (IGD) \cite{IGD}, normalized hypervolume (NHV) \cite{HypE}, pure diversity (PD) \cite{indicator-PD}, Spacing \cite{indicator-spacing}, and Spread ($\Delta$) \cite{spread-delta}.
\NAME{} also provides a lot of widely-used operators for different encodings \cite{SBX,poly-mut,operator-TSPCrossover,operator-TSPMutation,DE,PSO,FEP}, which can be used together with all the MOEAs in \NAME{}.

\item{\emph{Good Usability.}} \NAME{} is fully developed in MATLAB language, thus any machines installed with MATLAB can use \NAME{} regardless of the operating system.
Besides, users do not need to write any additional code when performing experiments using MOEAs in \NAME{}, as \NAME{} provides a user-friendly GUI, where users can configure all the settings and perform experiments on MOEAs via the GUI, and the experimental results can further be saved as a table in the format of Excel or LaTeX.
In other words, with the assistance of \NAME{}, users can directly obtain the statistical experimental results to be used in academic writings by \emph{one-click} operation via the GUI.

\item{\emph{Easy Extensibility.}} \NAME{} is not only easy to be used, but also easy to be extended.
To be specific, the source code of all the MOEAs, MOPs and operators in \NAME{} are completely open source, and the length of the source code is very short due to the advantages of matrix operation in MATLAB, such that users can easily implement their own MOEAs, MOPs and operators on the basis of existing resources in \NAME{}.
In addition, all new MOEAs developed on the basis of interfaces provided by \NAME{} can be also included into the platform, such that the library in \NAME{} can be iteratively updated by all users to follow state-of-the-arts.

\item{\emph{Delicate Considerations.}} There are many delicate considerations in the implementation of \NAME{}.
For example, \NAME{} provides different figure demonstrations of experimental results, and it also provides well-designed sampling methods for different shapes of Pareto optimal fronts.
Fig. \ref{fig:PF} shows the reference points sampled by \NAME{} on the Pareto optimal fronts of some MOPs with 3 objectives, while such reference points have not been provided by any other existing EMO libraries.
It is also worth noting that, since the efficiency of most MOEAs is subject to the non-dominated sorting process, \NAME{} employs the efficient non-dominated sort ENS-SS~\cite{ENS} for two-objective optimization and the tree-based ENS termed T-ENS~\cite{LMEA} for optimization with more than two objectives as the non-dominated sorting approaches, which have been demonstrated to be much more efficient than the fast non-dominated sort \cite{NSGA-II} used in other EMO libraries.

\end{itemize}

\begin{figure}
  \centering
  \begin{overpic}[width=0.8\linewidth]{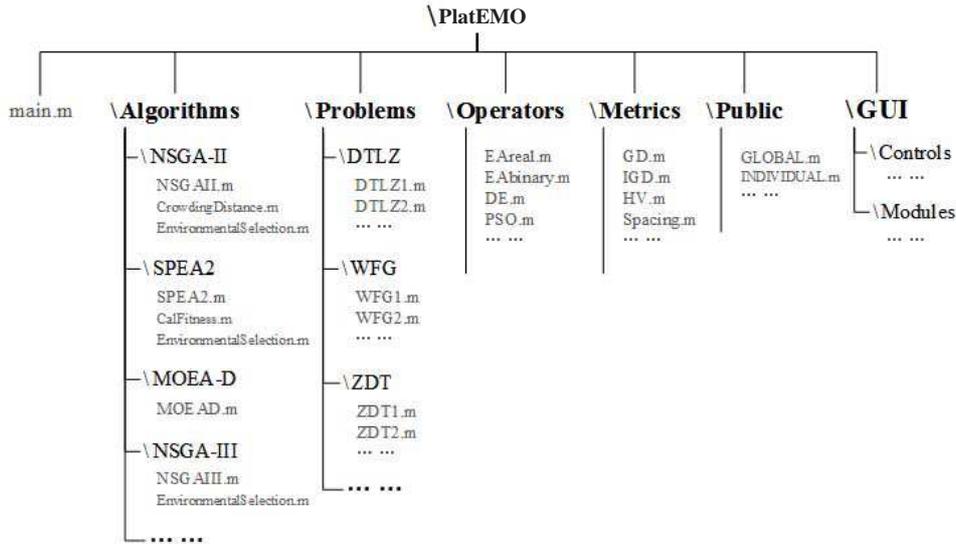}
  \put(45.5,56.5){\footnotesize{\textbf{\NAME{}}}}
  \end{overpic}
  \caption{Basic file structure of \NAME{}.}
  \label{fig:file}
\end{figure}

The rest of this paper is organized as follows. In the next section, the architecture of \NAME{} is presented on several aspects, i.e., the file structure of PlatEMO, the class diagram of PlatEMO, and the sequence diagram of executing algorithms by PlatEMO.
Section III introduces how to use \NAME{} for analyzing the performance of algorithms and performing comparative experiments.
The methods of extending \NAME{} with new MOEAs, MOPs, operators and performance indicators are described in Section IV.
Finally, conclusion and future work are given in Section V.

\section{Architecture of \NAME{}}

After opening the root directory of \NAME{}, users can see a lot of \emph{.m} files organized in the structure shown in Fig. \ref{fig:file}, where it is very easy to find the source code of specified MOEAs, MOPs, operators or performance indicators.
As shown in Fig. \ref{fig:file}, there are six folders and one interface function \emph{main.m} in the root directory of \NAME{}.
The first folder \emph{$\backslash$Algorithms} is used to store all the MOEAs in \NAME{}, where each MOEA has an independent subfolder and all the relevant functions are in it.
For instance, as shown in Fig. \ref{fig:file}, the subfolder \emph{$\backslash$Algorithms$\backslash$NSGA-II} contains three functions \emph{NSGAII.m}, \emph{CrowdingDistance.m} and \emph{EnvironmentalSelection.m}, which are used to perform the main loop, calculate the crowding distances, and perform the environmental selection of NSGA-II, respectively.
The second folder \emph{$\backslash$Problems} contains a lot of subfolders for storing benchmark MOPs.
For example, the subfolder \emph{$\backslash$Problems$\backslash$DTLZ} contains 14 DTLZ test problems (i.e., DTLZ1--DTLZ9, C1\_DTLZ1, C2\_DTLZ2, C3\_DTLZ4, IDTLZ1 and IDTLZ2), and the subfolder \emph{$\backslash$Problems$\backslash$WFG} contains 9 WFG test problems (i.e., WFG1--WFG9).
The folders \emph{$\backslash$Operators} and \emph{$\backslash$Metrics} store all the operators and performance indicators, respectively.
The next folder \emph{$\backslash$Public} is used to store some public classes and functions, such as \emph{GLOBAL.m} and \emph{INDIVIDUAL.m}, which are two classes in \NAME{} representing settings of parameters and definitions of individuals, respectively.
The last folder \emph{$\backslash$GUI} stores all the functions for establishing the GUI of \NAME{}, where users need not read or modify them.

\begin{figure}
  \centering
  \includegraphics[width=0.8\linewidth]{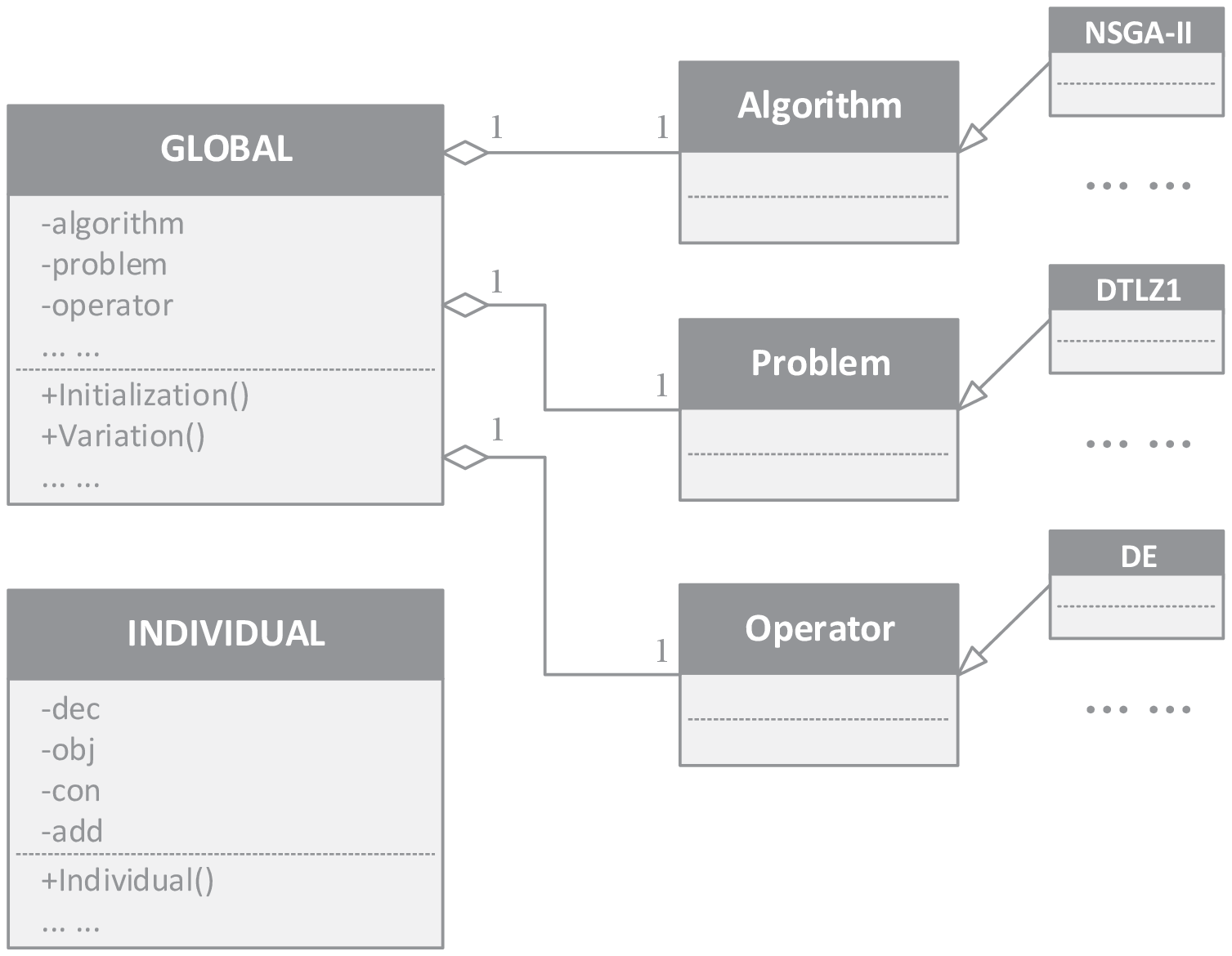}\\
  \caption{Class diagram of the architecture of \NAME{}.}
  \label{fig:class}
\end{figure}

\NAME{} also has a simple architecture, where it only involves two classes, namely \emph{GLOBAL} and \emph{INDIVIDUAL}, to store all the parameters and joint all the components (e.g., MOEAs, MOPs and operators).
The class diagram of these two classes is presented in Fig. \ref{fig:class}.
The first class \emph{GLOBAL} represents all the parameter setting, including the handle of MOEA function \emph{algorithm}, the handle of MOP function \emph{problem}, the handle of operator function \emph{operator} and other parameters about the environment (the population size, the number of objectives, the length of decision variables, the maximum number of fitness evaluations, etc.).
Note that all the properties in \emph{GLOBAL} are read-only, which can only be assigned by users when the object is being instantiated.
\emph{GLOBAL} also provides several methods to be invoked by MOEAs, where MOEAs can achieve some complex operations via these methods.
For instance, the method \emph{Initialization}() can generate a randomly initial population with specified size, and the method \emph{Variation}() can generate a set of offsprings with specified parents.

%When invoking a method of \emph{GLOBAL}, MOEAs need not know about the MOP function or the operator function, since all the configurations have been stored in the \emph{GLOBAL} object.

The other class in \NAME{} is \emph{INDIVIDUAL}, where its objects are exactly individuals in MOEAs.
An \emph{INDIVIDUAL} object contains four properties, i.e., \emph{dec}, \emph{obj}, \emph{con} and \emph{add}, all of which are also read-only.
\emph{dec} is the array of decision variables of the individual, which is assigned when the object is being instantiated.
\emph{obj} and \emph{con} store the objective values and the constraint values of the individual, respectively, which are calculated after \emph{dec} has been assigned.
The property \emph{add} is used to store additional properties of the individual for some special operators, such as the 'speed' property in PSO operator \cite{PSO}.

\begin{figure}
  \centering
  \includegraphics[width=0.9\linewidth]{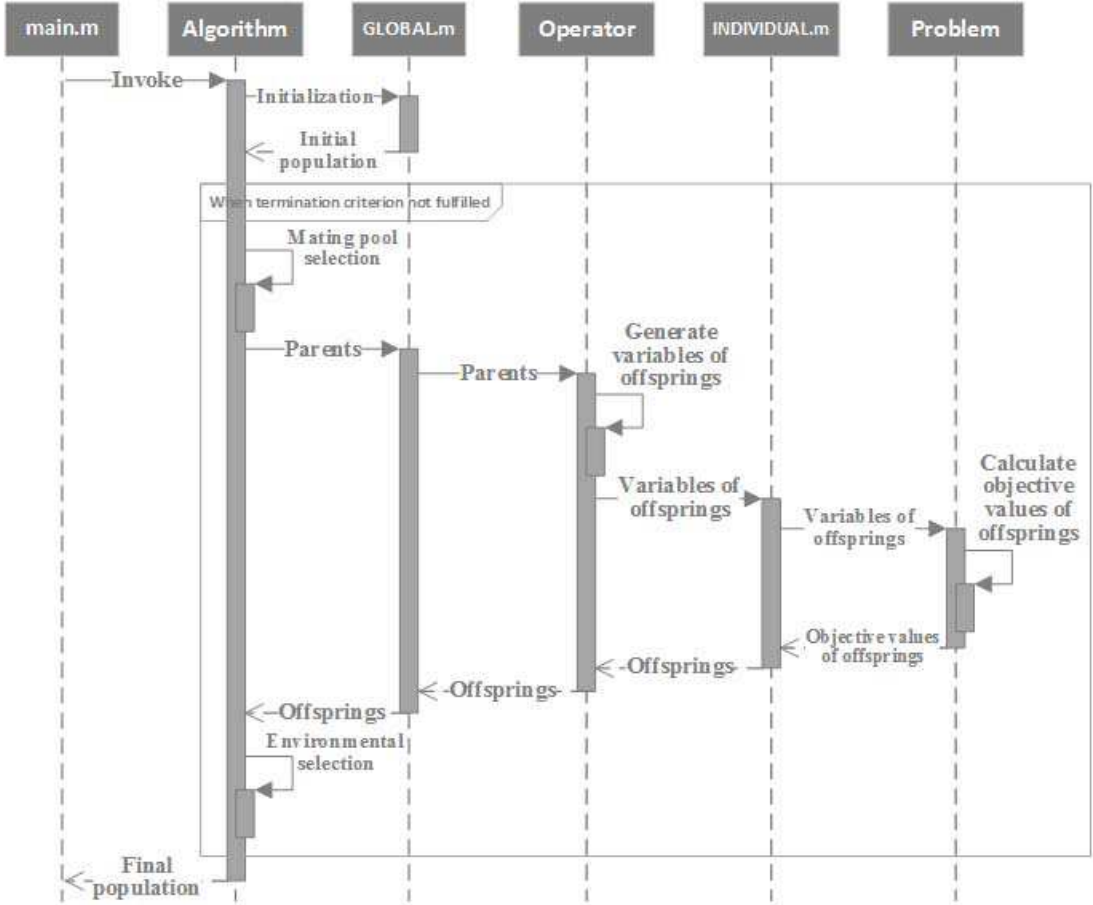}\\
  \caption{Sequence diagram of running a general multi-objective optimization algorithm by \NAME{} without GUI.}
  \label{fig:sequence}
\end{figure}

In order to better understand the mechanism of \NAME{}, Fig. \ref{fig:sequence} illustrates the sequence diagram of running an MOEA by \NAME{} without GUI.
To begin with, the interface \emph{main.m} first invokes the algorithm function (e.g., \emph{NSGAII.m}), then the algorithm obtains an initial population (i.e., an array of \emph{INDIVIDUAL} objects) from the \emph{GLOBAL} object by invoking its method \emph{Initialization}().
After that, the algorithm starts the evolution until the termination criterion is fulfilled, where the maximum number of fitness evaluations is used as the termination criterion for all the MOEAs in \NAME{}.
In each generation of a general MOEA, it first performs mating pool selection for selecting a number of parents from the current population, and the parents are used to generate offsprings by invoking the method \emph{Variation}() of \emph{GLOBAL} object.
\emph{Variation}() then passes the parents to the operator function (e.g., \emph{DE.m}), which is used to calculate the decision variables of the offsprings according to the parents.
Next, the operator function invokes the \emph{INDIVIDUAL} class to instantiate the offspring objects, where the objective values of offsprings are calculated by invoking the problem function (e.g., \emph{DTLZ1.m}).
After obtaining the offsprings, the algorithm performs environmental selection on the current population and the offsprings to select the population for next generation.
When the number of instantiated \emph{INDIVIDUAL} objects exceeds the maximum number of fitness evaluations, the algorithm will be terminated and the final population will be output.

As presented by the above procedure, the algorithm function, the problem function and the operator function do not invoke each other directly; instead, they are connected to each other by the \emph{GLOBAL} class and the \emph{INDIVIDUAL} class.
This mechanism has two advantages. First, MOEAs, MOPs and operators in \NAME{} are independent mutually, and they can be arbitrarily combined with each other, thus providing high flexibility \NAME{}.
Second, users need not consider the details of the MOP or the operator to be involved when developing a new MOEA, thus significantly improving the development efficiency.

\section{Running \NAME{}}

As mentioned in Section I, \NAME{} provides two ways to run it: first, it can be run with a GUI by invoking the interface \emph{main}() without input parameter, then users can perform MOEAs on MOPs by simple \emph{one-click} operations; second, it can be run without GUI, and users can perform one MOEA on an MOP by invoking \emph{main}() with input parameters.
In this section, we elaborate these two ways of running \NAME{}.

\subsection{Running \NAME{} without GUI}

\begin{table}[t]
\renewcommand{\arraystretch}{0.8}
\caption{All the Acceptable Parameters for the Interface of \NAME{}.}
\begin{center}
\begin{tabular}{|c|c|c|c|}
\hline
Parameter&\MR{2}{Type}&Default&\MR{2}{Description}\\
Name&&Value&\\
\hline
\emph{-algorithm}&function handle&\scriptsize{@NSGAII}&Algorithm function\\
\hline
\emph{-problem}&function handle&\scriptsize{@DTLZ2}&Problem function\\
\hline
\emph{-operator}&function handle&\scriptsize{@EAreal}&Operator function\\
\hline
\emph{-N}&positive integer&100&Population size\\
\hline
\MR{2}{\emph{-M}}&\MR{2}{positive integer}&\MR{2}{3}&Number of\\
&&&objectives\\
\hline
\MR{2}{\emph{-D}}&\MR{2}{positive integer}&\MR{2}{12}&Number of\\
&&&decision variables\\
\hline
\MR{3}{\emph{-evaluation}}&\MR{3}{positive integer}&\MR{3}{10000}&Maximum number\\
&&&of fitness\\
&&&evaluations\\
\hline
\emph{-run}&positive integer&1&Run No.\\
\hline
\MR{3}{\emph{-mode}}&\MR{3}{1 or 2}&\MR{3}{1}&Run mode\\
&&&(1. display result)\\
&&&(2. save result)\\
\hline
\MR{3}{\emph{-X\_parameter}}&\MR{3}{cell}&\MR{3}{N/A}&The parameter\\
&&&values for\\
&&&function X\\
\hline
\end{tabular}
\end{center}
\label{tab:Parameter}
\end{table}

The interface \emph{main}() can be invoked with a set of input parameters by the following form: \emph{main}(\emph{'name1', value1, 'name2', value2, ...}), where \emph{name1, name2, ...} denote the names of the parameters and \emph{value1, value2, ...} denote the values of the parameters. All the acceptable parameters together with their data types and default values are listed in Table \ref{tab:Parameter}.
It is noteworthy that every parameter has a default value such that users need not assign all the parameters.
As an example, the command \emph{main}(\emph{'-algorithm',@NSGAII,'-problem',@DTLZ2,'-N',100,'-M',3,'-D',10,'-evaluation',10000}) is used to perform NSGA-II on DTLZ2 with a population size of 100, an objective number of 3, a decision variable length of 10, and a maximum fitness evaluation number of 10000.

\begin{figure}
  \centering
  \includegraphics[width=0.6\linewidth]{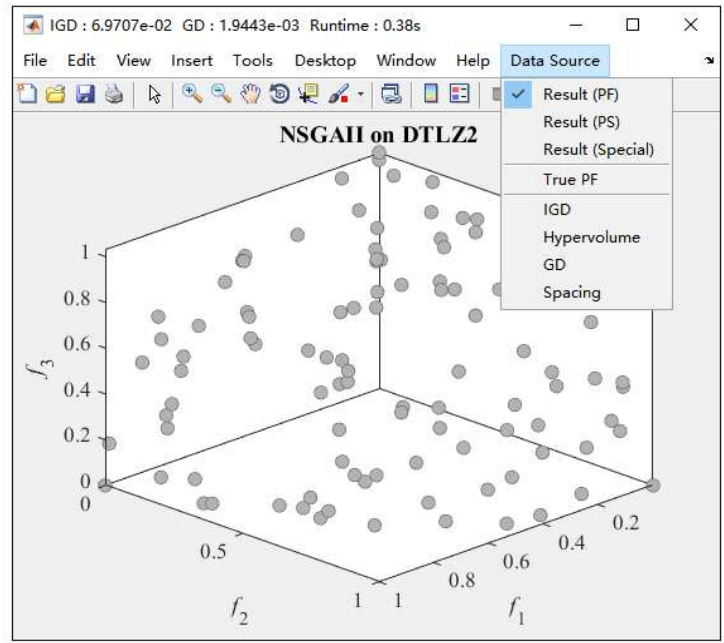}\\
  \caption{The objective values of the population obtained by NSGA-II on DTLZ2 with 3 objectives by running \NAME{} without GUI.}
  \label{fig:image1}
\end{figure}

By invoking \emph{main}() with parameters, one MOEA can be performed on an MOP with the specified setting, while the GUI will not be displayed. After the MOEA has been terminated, the final population will be displayed or saved, which is determined by the parameter \emph{-mode} shown in Table \ref{tab:Parameter}.
To be specific, if \emph{-mode} is set to 1, the objective values or decision variable values of the final population will be displayed in a new figure, and users can also observe the true Pareto front and the evolutionary trajectories of performance indicator values. For example, Fig. \ref{fig:image1} shows the objective values of the population obtained by NSGA-II on DTLZ2 with 3 objectives, where users can select the figure to be displayed on the rightmost menu. If \emph{-mode} is set to 2, the final population will be saved in a \emph{.mat} file, while no figure will be displayed.

Generally, there are four parameters to be assigned by users as listed in Table \ref{tab:Parameter} (i.e., the population size \emph{-N}, the number of objectives \emph{-M}, the number of decision variables \emph{-D}, and the maximum number of fitness evaluations \emph{-evaluation}); however, different MOEAs, MOPs or operators may involve additional parameter settings.
For instance, there is a parameter $rate$ denoting the ratio of selected knee points in \emph{KnEA} \cite{KnEA}, and there are four parameters $proC$, $disC$, $proM$ and $disM$ in \emph{EAreal} \cite{SBX,poly-mut}, which denote the crossover probability, the distribution index of simulated binary crossover, the number of bits undergone mutation, and the distribution index of polynomial mutation, respectively.
In \NAME{}, such function related parameters can also be assigned by users via assigning the parameter \emph{-X\_parameter}, where \emph{X} indicates the name of the function.
For example, users can use the command \emph{main}(\emph{\ldots,'-KnEA\_parameter',\{0.5\},\ldots}) to set the value of $rate$ to 0.5 for \emph{KnEA}, and use the command \emph{main}(\emph{\ldots,'-EAreal\_parameter',\{1,20,1,20\},\ldots}) to set the values of $proC$, $disC$, $proM$ and $disM$ to 1, 20, 1 and 20 for \emph{EAreal}, respectively.
Besides, users can find the acceptable parameters of each MOEA, MOP and operator in the comments at the beginning of the related function.

%Invoking \emph{main}() with parameters provides a simple way for users to run \NAME{}, but users need to process the results for further studies by themselves. By contrast, the GUI of \NAME{} can process the results automatically.

\subsection{Running \NAME{} with GUI}

\begin{figure*}
  \centering
  \includegraphics[width=1\linewidth]{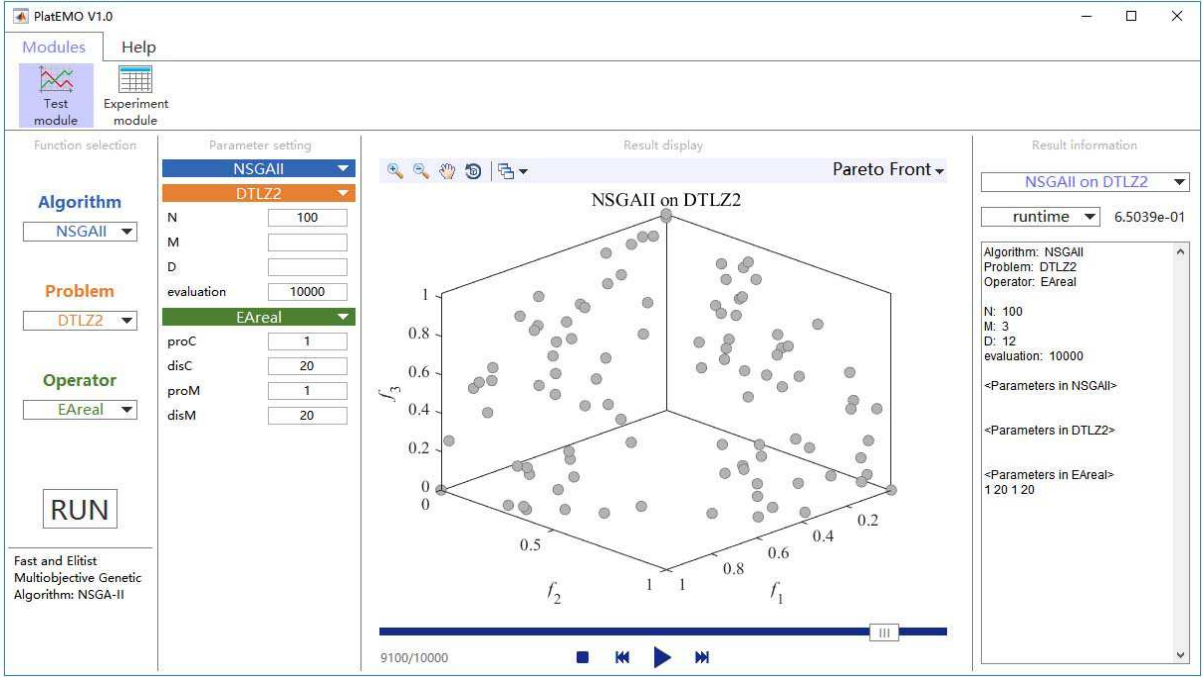}\\
  \caption{The test module of \NAME{}.}
  \label{fig:image2}
\end{figure*}

The GUI of \NAME{} currently contains two modules. The first module is used to analyze the performance of each MOEA, where one MOEA on an MOP can be performed in this module each time, and users can observe the result via different figure demonstrations.
The second module is designed for statistical experiments, where multiple MOEAs on a batch of MOPs can be performed at the same time, and the statistical experimental results can be saved as Excel table or LaTeX table.

The interface of the first module, i.e., test module, is shown in Fig. \ref{fig:image2}. As can be seen from the figure, the main panel is divided into four parts. The first subpanel from left provides three popup menus, where users can select the MOEA, MOP and operator to be performed. The second subpanel lists all the parameters to be assigned, which depends on the selected MOEA, MOP and operator. The third subpanel displays the current population during the optimization, other figures such as the true Pareto front and the evolutionary trajectories of performance indicator values can also be displayed. In addition, users can observe the populations in previous generations by dragging the slider at the bottom. The fourth subpanel stores the detailed information of historical results.
As a result, the test module provides similar functions to the \NAME{} without GUI, but users do not need to write any additional command or code when using it.

\begin{figure*}
  \centering
  \includegraphics[width=1\linewidth]{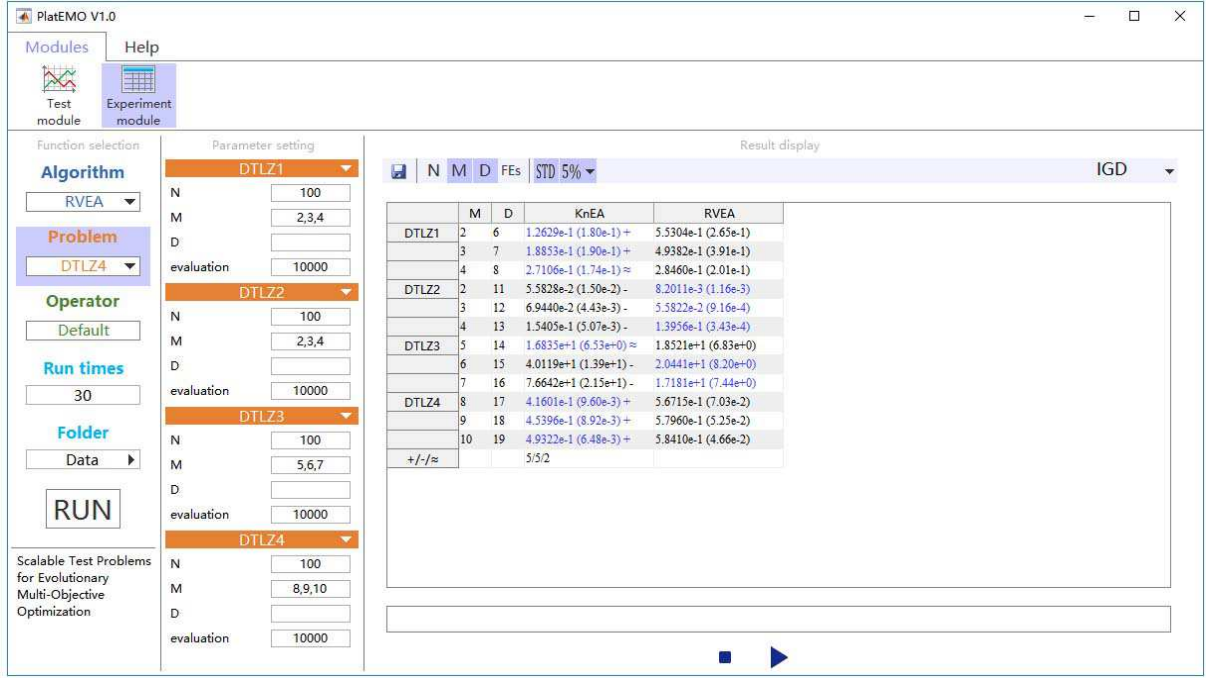}\\
  \caption{The experimental module of \NAME{}.}
  \label{fig:image3}
\end{figure*}

The other module on the GUI is the experimental module, which is shown in Fig. \ref{fig:image3}.
Similar to the text module, users should first select the MOEAs, MOPs and operators to be performed in the leftmost subpanel. Note that multiple MOEAs and MOPs can be selected in the experimental module.
After setting the number of total runs, folder for saving results, and all the relevant parameters, the experiment can be started and the statistical results will be shown in the rightmost subpanel.
Users can select any performance indicator to calculate the results to be listed in the table, where the mean and the standard deviation of the performance indicator value are shown in each grid.
Furthermore, the best result in each row is shown in blue, and the Wilcoxon rank sum test result is labeled by the signs '$+$', '$-$' and '$\approx$', which indicate that the result is significantly better, significantly worst and statistically similar to the result in the control column, respectively.
After the experiment is finished, the data shown in the table can be saved as Excel table (\emph{.xlsx} file) or LaTeX table (\emph{.tex} file).
For example, after obtaining the experimental results shown in the table in Fig. \ref{fig:image3}, users can press the 'saving' button on the GUI to save the table in the format of LaTeX, where the corresponding LaTeX table is shown in Table \ref{tab:example}.

\begin{table}[t]
\renewcommand{\arraystretch}{0.8}
\centering
\caption{IGD Values of KnEA and RVEA on DTLZ1--DTLZ4. The LaTeX Code of This Table is Automatically Generated by \NAME{}.}
\begin{tabular}{ccccc}
\toprule
Problem&$M$&$D$&KnEA&RVEA\\
\midrule
\multirow{3}{*}{DTLZ1}&2&6&\hl{1.2629e-1 (1.80e-1) $+$}&5.5304e-1 (2.65e-1)\\
&3&7&\hl{1.8853e-1 (1.90e-1) $+$}&4.9382e-1 (3.91e-1)\\
&4&8&\hl{2.7106e-1 (1.74e-1) $\approx$}&2.8460e-1 (2.01e-1)\\
\hline
\multirow{3}{*}{DTLZ2}&2&11&5.5828e-2 (1.50e-2) $-$&\hl{8.2011e-3 (1.16e-3)}\\
&3&12&6.9440e-2 (4.43e-3) $-$&\hl{5.5822e-2 (9.16e-4)}\\
&4&13&1.5405e-1 (5.07e-3) $-$&\hl{1.3956e-1 (3.43e-4)}\\
\hline
\multirow{3}{*}{DTLZ3}&5&14&\hl{1.6835e+1 (6.53e+0) $\approx$}&1.8521e+1 (6.83e+0)\\
&6&15&4.0119e+1 (1.39e+1) $-$&\hl{2.0441e+1 (8.20e+0)}\\
&7&16&7.6642e+1 (2.15e+1) $-$&\hl{1.7181e+1 (7.44e+0)}\\
\hline
\multirow{3}{*}{DTLZ4}&8&17&\hl{4.1601e-1 (9.60e-3) $+$}&5.6715e-1 (7.03e-2)\\
&9&18&\hl{4.5396e-1 (8.92e-3) $+$}&5.7960e-1 (5.25e-2)\\
&10&19&\hl{4.9322e-1 (6.48e-3) $+$}&5.8410e-1 (4.66e-2)\\
\hline
\multicolumn{3}{c}{$+/-/\approx$}&5/5/2&\\
\bottomrule
\end{tabular}
\label{tab:example}
\end{table}

It can be concluded from the above introduction that the functions provided by \NAME{} are modularized, where two modules (i.e., the test module and the experimental module) are included in the current version of \NAME{}. In the future, we also plan to develop more modules to provide more functions for users.

\section{Extending \NAME{}}

\NAME{} is an open platform for scientific research and applications of EMO, hence it allows users to add their own MOEAs, MOPs, operators and performance indicators to it, where users should save the new MOEA, MOP, operator or performance indicator to be added as a MATLAB function (i.e., a \emph{.m} file) with the specified interface and form, and put it in the corresponding folder.
In the following, the methods of extending \NAME{} with a new MOEA, MOP, operator and performance indicator are illustrated by several cases, respectively.

\subsection{Adding New Algorithms to \NAME{}}

\begin{figure}
  \centering
  \includegraphics[width=0.7\linewidth]{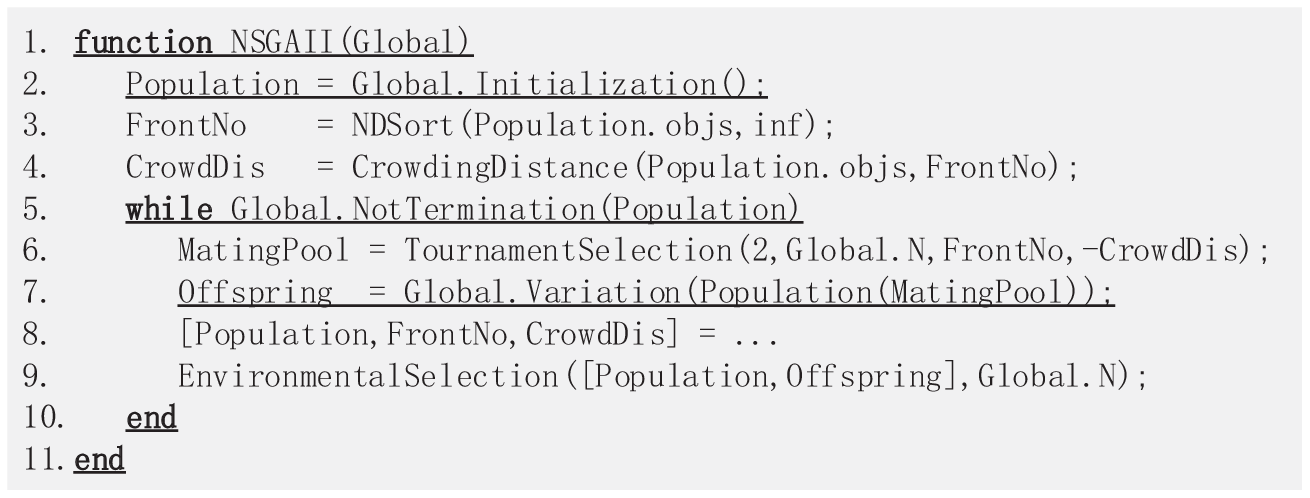}\\
  \caption{The source code of the main function of NSGA-II. The common code required by any MOEA is underlined.}
  \label{fig:code1}
\end{figure}

All the \emph{.m} files of MOEA functions are stored in the folder \emph{$\backslash$Algorithms} in the root directory of \NAME{}, and all the \emph{.m} files for the same MOEA should be put in the same subfolder.
For example, as shown in the file structure in Fig. \ref{fig:file}, the three \emph{.m} files for NSGA-II (i.e., \emph{NSGAII.m}, \emph{CrowdingDistance.m} and \emph{EnvironmentalSelection.m}) are all in the subfolder \emph{$\backslash$Algorithms$\backslash$NSGA-II}.
A case study including the source code of the main function of NSGA-II (i.e. \emph{NSGAII.m}) is given in Fig. \ref{fig:code1}, where the logic of the function is completely the same to the one shown in Fig. \ref{fig:sequence}.

To begin with, the main function of an MOEA has one input parameter and zero output parameter, where the only input parameter denotes the \emph{GLOBAL} object for the current run.
Then an initial population \emph{Population} is generated by invoking \emph{Global.Initialization}(), and the non-dominated front number and the crowding distance of each individual are calculated (line 2--4).
In each generation, the function \emph{Global.NotTermination}() is invoked to check whether the termination criterion is fulfilled, and the variable \emph{Population} is passed to this function to be the final output (line 5).
Afterwards, the mating pool selection, generating offsprings, and environmental selection are performed in sequence (line 6--9).

The common code required by any MOEA is underlined in Fig. \ref{fig:code1}.
In addition to the interface of the function, one MOEA needs to perform at least the following three operations: obtaining an initial population via \emph{Global.Initialization}(), checking the optimization state and outputting the final population via \emph{Global.NotTermination}(), and generating offsprings via \emph{Global.Variation}(), where all these three functions are provided by the \emph{GLOBAL} object.
Apart from the above three common operations, different MOEAs may have different logics and different functions to be invoked.

\subsection{Adding New Problems to \NAME{}}

\begin{figure}
  \centering
  \includegraphics[width=0.7\linewidth]{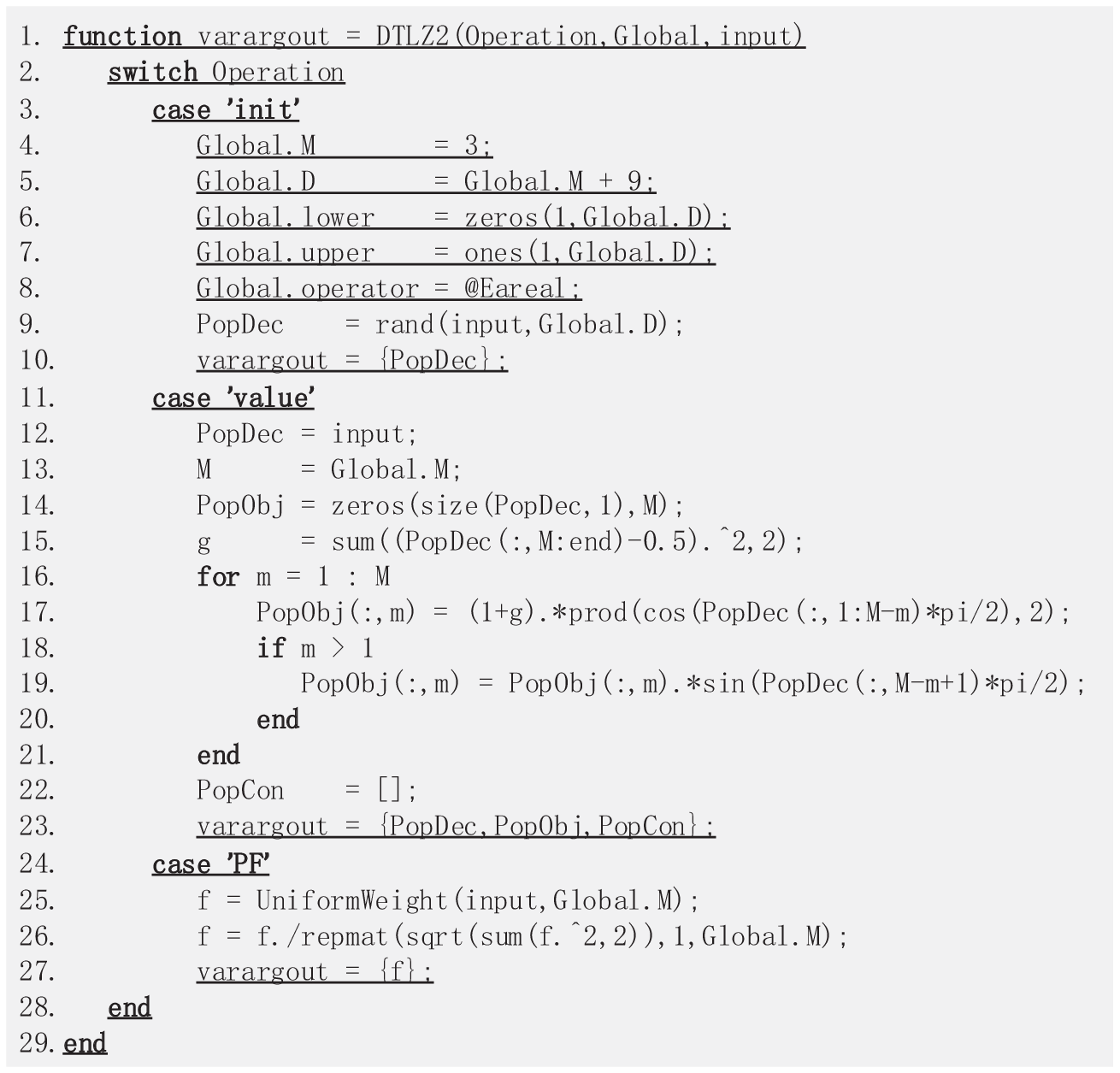}\\
  \caption{The source code of DTLZ2. The common code required by any MOP is underlined.}
  \label{fig:code2}
\end{figure}

All the \emph{.m} files of MOP functions are stored in the folder \emph{$\backslash$Problems}, and one \emph{.m} file usually indicates one MOP.
Fig. \ref{fig:code2} gives the source code of DTLZ2, where the common code required by any MOP is underlined.
It can be seen from the source code that, the interface of DTLZ2 is more complex than the one of NSGA-II, where the function \emph{DTLZ2}() includes three input parameters and one output parameter.
The input parameter \emph{Operation} determines the operation to be performed; the parameter \emph{Global} denotes the \emph{GLOBAL} object; and the parameter \emph{input} has different meanings when \emph{Operation} is set to different values, so does the output parameter \emph{varargout}.

Different from the MOEA functions which are invoked only once in each run, an MOP function may be invoked many times for different operations.
As shown in Fig. \ref{fig:code2}, an MOP function contains three independent operations: generating random decision variables (line 3--10), calculating objective values and constraint values (line 11--23), and sampling reference points on the true Pareto front (line 24--27).
To be specific, if \emph{Operation} is set to \emph{'init'}, the MOP function will return the decision variables of a random population with size \emph{input} (line 9--10).
Meanwhile, it sets \emph{Global.M}, \emph{Global.D}, \emph{Global.lower}, \emph{Global.upper} and \emph{Global.operator} to their default values, which denote the number of objectives, number of decision variables, lower boundary of each decision variable, upper boundary of each decision variable, and the operator function, respectively (line 4--8).
When \emph{Operation} is set to \emph{'value'}, the parameter \emph{input} will denote the decision variables of a population, and the objective values and constraint values of the population will be calculated and returned according to the decision variables (line 14--23).
And if \emph{Operation} is set to \emph{'PF'}, a number of \emph{input} uniformly distributed reference points will be sampled on the true Pareto front and returned (line 25--27).

\subsection{Adding New Operators or Performance Indicators to \NAME{}}

\begin{figure}
  \centering
  \includegraphics[width=0.7\linewidth]{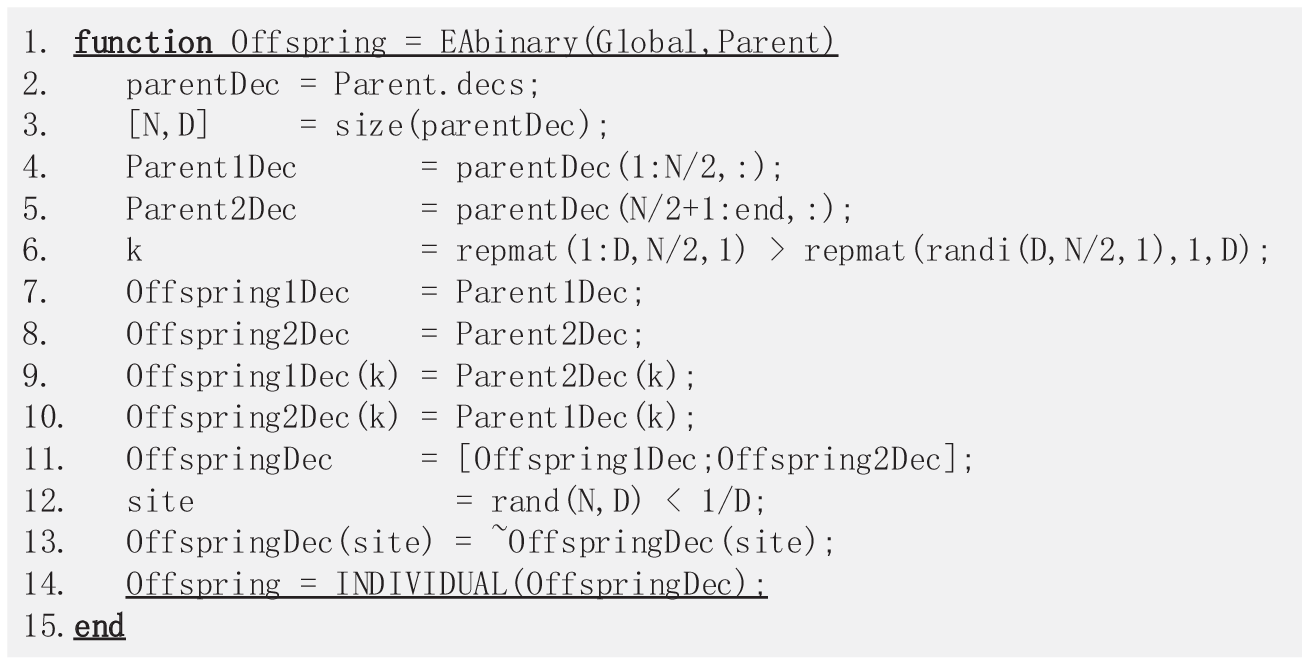}\\
  \caption{The source code of evolutionary operator based on binary coding. The common code required by any operator is underlined.}
  \label{fig:code3}
\end{figure}

Fig. \ref{fig:code3} shows the source code of evolutionary operator based on binary coding (i.e. \emph{EAbinary.m}), where the \emph{.m} files of the operator functions are all stored in the folder \emph{$\backslash$Operators}.
An operator function has two input parameters, one denoting the \emph{GLOBAL} object (i.e. \emph{Global}) and the other denoting the parent population (i.e. \emph{Parent}), and it also has one output parameter denoting the generated offsprings (i.e. \emph{Offspring}).
As can be seen from the source code in Fig. \ref{fig:code3}, the main task of an operator function is to generate offsprings according to the values of \emph{Parent}, where \emph{EAbinary}() performs the single-point crossover in line 6--11 and the bitwise mutation in line 12--13 of the code. Afterwards, the \emph{INDIVIDUAL} objects of the offsprings are generated and returned (line 14).

\begin{figure}
  \centering
  \includegraphics[width=0.7\linewidth]{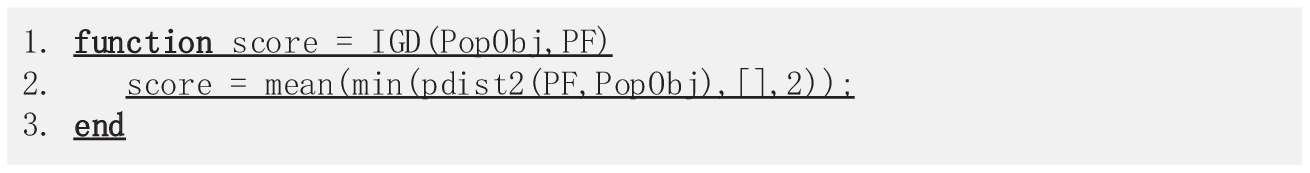}\\
  \caption{The source code of IGD. The common code required by any performance indicator is underlined.}
  \label{fig:code4}
\end{figure}

Fig. \ref{fig:code4} shows the source code of IGD, where all these performance indicator functions are stored in the folder \emph{$\backslash$Metrics}.
The task of a performance indicator is to calculate the indicator value of a population according to a set of reference points. The input parameters of \emph{IGD}() consists of two parts: the objective values of the population (i.e. \emph{PopObj}), and the reference points sampled on the true Pareto front (i.e. \emph{PF}). Correspondingly, the output parameter of \emph{IGD}() is the IGD value (i.e. \emph{score}).
Thanks to the merits of matrix operation in MATLAB, the source code of IGD is quite short as shown in Fig. \ref{fig:code4}, where the calculation of the mean value of the minimal distance of each point in \emph{PF} to the points in \emph{PopObj} can be performed using a built-in function \emph{pdist2}() provided by MATLAB.

\subsection{Adding Acceptable Parameters for New Functions}

\begin{figure}
  \centering
  \includegraphics[width=0.7\linewidth]{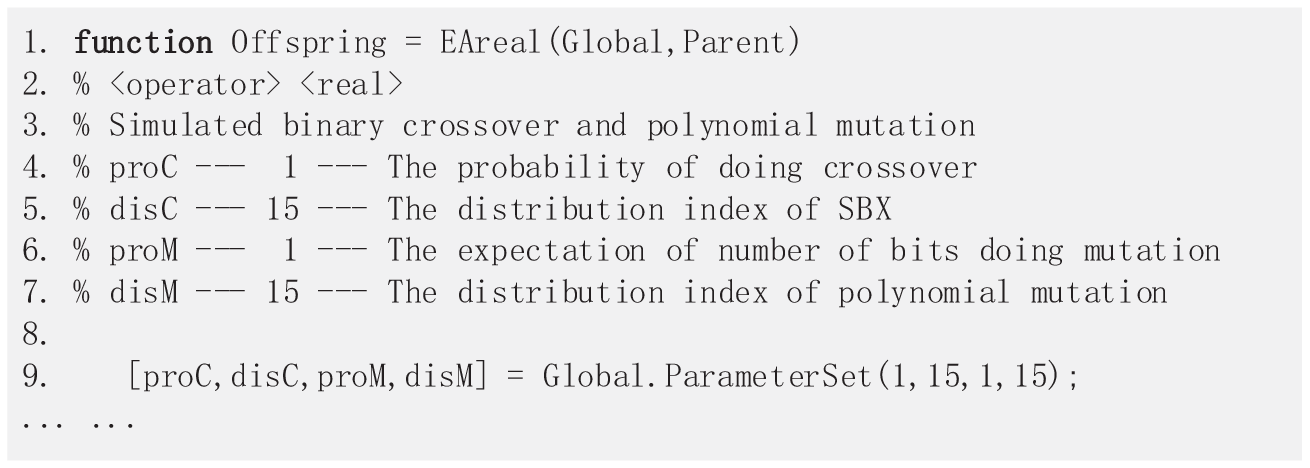}\\
  \caption{The comments and the source code in the head of the function of evolutionary operator based on real value coding.}
  \label{fig:code5}
\end{figure}

All the user-defined functions can have their own parameters as well as the functions provided by \NAME{}, where these parameters can be either assigned by invoking \emph{main}(\emph{\ldots,'-X\_parameter',\{\ldots\},\ldots}) with \emph{X} denoting the function name, or displayed on the GUI for assignment.
In order to add acceptable parameters for an MOEA, MOP, operator or performance indicator function, the comments in the head of the function should be written in a specified form.
To be specific, Fig. \ref{fig:code5} shows the comments and the source code in the head of the function of evolutionary operator based on real value coding (i.e. \emph{EAreal.m}).

The comment in line 2 of Fig. \ref{fig:code5} gives the two labels of this function, which are used to make sure this function can be identified by the GUI.
The comment in line 3 is a brief introduction about this function; for an MOEA or MOP function, such introduction should be the title of the relevant literature.
The parameters $proC$, $disC$, $proM$ and $disM$ for this function are given by the comments in line 4--7, where the names of the parameters are in the first column, the default values of the parameters are in the second column, and the introductions about the parameters are given in the third column.
The columns in each row are divided by the sign '---'.

The comments define the parameters and their default values for the function, and invoking \emph{Global.ParameterSet}() can make these parameters assignable to users.
As shown in line 9 of Fig. \ref{fig:code5}, the function invokes \emph{Global.ParameterSet}() with four inputs denoting the default values of the parameters, and sets the four parameters to the outputs.
More specifically, if users have not assigned the parameters, they will equal to their default values (i.e. 1, 15, 1 and 15).
Otherwise, if users assign the parameters by invoking \emph{main}(\emph{\ldots,'-EAreal\_parameter',\{a,b,c,d\},\ldots}), the parameters $proC$, $disC$, $proM$ and $disM$ will be set to \emph{a}, \emph{b}, \emph{c} and \emph{d}, respectively.

\section{Conclusion and Future Work}

This paper has introduced a MATLAB-based open source platform for evolutionary multi-objective optimization, namely \NAME{}.
The current version of \NAME{} includes 50 multi-objective optimization algorithms and 110 multi-objective test problems, having covered the majority of state-of-the-arts.
Since \NAME{} is developed on the basis of a light architecture with simple relations between objects, it is very easy to be used and extended.
Moreover, \NAME{} provides a user-friendly GUI with a powerful experimental module, where engineers and researchers can use it to quickly perform their experiments without writing any additional code.

This paper has described the architecture of \NAME{}, and it has also introduced the steps of running \NAME{} with and without the GUI.
Then, the ways of adding new algorithms, problems, operators and performance indicators to \NAME{} have been elaborated by several cases.

We will continuously maintain and develop \NAME{} in the future. On one hand, we will keep following the state-of-the-arts and adding more effective algorithms and new problems into \NAME{}. On the other hand, more modules will be developed to provide more functions for users, such as preference optimization, dynamic optimization, noisy optimization, etc.
We hope that \NAME{} is helpful to the researchers working on evolutionary multi-objective optimization, and we sincerely encourage peers to join us to shape the platform for better functionality and usability.

\section*{Ackonwledgement}
This work was supported in part by National Natural Science Foundation of China (Grant No. 61672033, 61272152, 615012004, 61502001), and the Joint Research Fund for Overseas Chinese, Hong Kong and Macao Scholars of the National Natural Science Foundation of China (Grant No. 61428302). This manuscript was written during Y. Tian's visit at the Department of Computer Science, University of Surrey.
The authors would like to thank Mr. Kefei Zhou and Mr. Ran Xu for their valued work in testing the PlatEMO.

%\bibliographystyle{IEEEtran}
%\bibliography{references}

% Generated by IEEEtran.bst, version: 1.13 (2008/09/30)

\end{document}